# Neighborhood Averaging for Improving Outlier Detectors

*Jiawei Yang [a], Susanto Rahardja [ab]\*, Pasi Fränti [c]*

[a] Northwestern Polytechnical University
[b] Singapore Institute of Technology
[c] University of Eastern Finland

*Abstract —* **We hypothesize that *similar objects should have similar outlier scores*. To our knowledge, all existing outlier detectors calculate the outlier score for each object independently regardless of the outlier scores of the other objects. Therefore, they do not guarantee that similar objects have similar outlier scores. To verify our proposed hypothesis, we propose an outlier score post-processing technique for outlier detectors, called neighborhood averaging (NA), which pays attention to objects and their neighbors and guarantees them to have more similar outlier scores than their original scores. Given an object and its outlier score from any outlier detector, NA modifies its outlier score by combining it with its *k* nearest neighbors' scores. We demonstrate the effectivity of NA by using the well-known *k*-nearest neighbors (*k*-NN). Experimental results show that NA improves all 10 tested baseline detectors by 13% (from 0.70 to 0.79 AUC) on average evaluated on nine real-world datasets. Moreover, even outlier detectors that are already based on *k*-NN are also improved. The experiments also show that in some applications, the choice of detector is no more significant when detectors are jointly used with NA, which may pose a challenge to the generally considered idea that the data model is the most important factor. We open our code on www.outlierNet.com for reproducibility.**

## I. INTRODUCTION

Outliers are objects that significantly deviate from other objects. Outliers can indicate useful information, which can be applied in applications such as fraud detection [1, 2], abnormal time series [3, 4], and traffic patterns [5, 6]. Outliers can also be harmful because they are generally unwanted, can be considered *errors*, and may have biased statistical analysis for applications like clustering [7, 8]. Recently, outlier detection has also been applied to manufacturing data [9] and industrial applications [10]. For these reasons, outliers need to be detected.

Most outlier detectors calculate the so-called *outlier score* for every object independently and then calculate the threshold scores that deviate significantly from the others and label them as outliers [11]. To improve the results of baseline outlier detectors, *ensemble techniques* have been developed to combine the outcomes of multiple detectors to obtain a more accurate detector. An example is the *average ensemble* [1], which calculates the average outlier score from multiple baseline detectors. This can potentially improve outlier detection by smoothing the result of a weak detector and placing more emphasis on those that agree on an individual object. However, the existing ensemble techniques merely use more detectors but do not attempt to ensemble outlier scores of neighboring objects. Their success is also bounded by the reliability of the baseline detectors.

The *outlier score* is a fundamental concept in all score-based outlier detectors. All outlier detectors assume that outlier objects should have significantly higher or lower outlier scores [1]. Except for that, no attention has been paid to the relationship between objects and their outlier scores. Because outlier objects are directly decided by their outlier scores, it is vital to understand their relationship. In this paper, we address this problem.

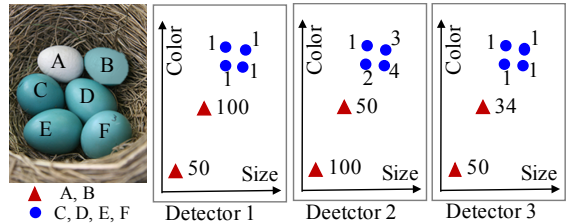

Fig. 1. Outlier scores given by three detectors on the task of detecting outlier eggs from a Robin. The results of Detector 3 can be obtained from the results of Detector 2 using the proposed method as shown in Fig. 7.

\* Corresponding author
{jiaweiyang,susantorahardja}@ieee.org; pasi.franti@uef.fi

In Fig. 1, all detectors successfully assign significantly higher scores to the outlier eggs (red triangles) but cannot guide the selection of the best detectors. We can see that egg A is distinctive and has the highest score. Detector 2 and Detector 3 are therefore better than Detector 1. Similarly, because eggs C, D, E, and F have the same color and size, they should have the same outlier scores. In this case, Detector 3 is better than Detector 2. Therefore, we can conclude that Detector 3 is the best among the three by comparing the objects' similarities.

Based on the case in Figure 1, we conclude that *similar objects should have similar outlier scores*. Although this could be seen as obvious, none of the state-of-the-art outlier detectors uses this. Many detectors simply make use of the objects' neighborhood in the process (especially all *k*-NN-based detectors), but they do not consider the result of the scores. For example, object B in Figure 1 has a high outlier score whereas all the objects near it have low scores. It should therefore have a lower score than object A, which has no normal objects in its vicinity.

To address the problem, we propose a novel *neighborhood averaging* (NA) technique. It post-processes the outlier scores provided by any existing outlier detector by averaging it with the scores of its neighbors. In other words, if an object is an outlier, it is more likely that its near neighbors are also outliers. In this case, the predicted score is enhanced. On the contrary, if the neighboring objects have low outlier scores (predicted as normalities), the score of the object is also reduced accordingly.

The beauty of NA is that it can serve as an additional and independent post-processing technique. It is different from ensemble techniques because rather than operating the results of multiple detectors of a single object, NA operates the results of multiple objects of a single detector as shown in Fig. 2. Neighborhood averaging is conceptually and fundamentally different from the ensemble techniques. It is also complementary to the ensembles, and these two approaches can be used jointly. While ensembles cannot assure similar objects have similar outlier scores, NA can achieve this.

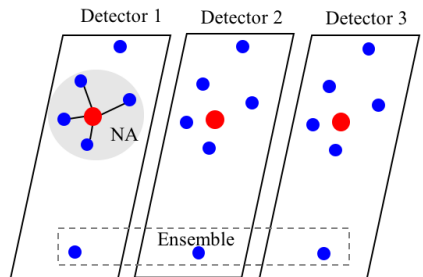

Fig. 2. Difference between NA and ensembles. Ensembles use multiple detectors' prediction of the *same* object (on the bottom), while NA uses a single detector's prediction of the *different* (neighboring) objects (with gray background).

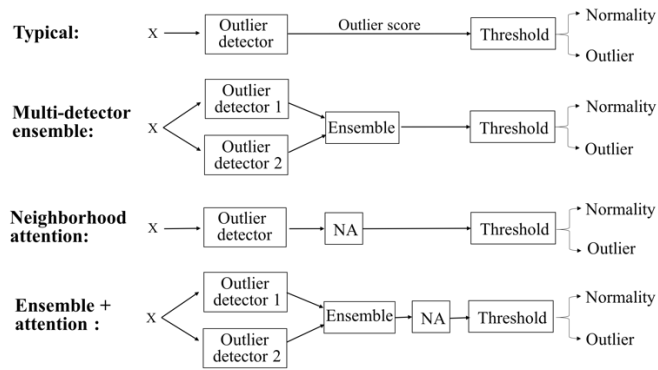

Fig. 3. Outlier detection process.

Fig. 3 demonstrates all the combinations that can be constructed from NA and the existing outlier detection methods, including ensemble techniques. On the top, we have the typical situation where dataset X is input into an outlier detector, which produces scores that are further processed by a threshold component to determine outliers. The second case is the multi-detector ensemble where the dataset is input into two outlier detectors to produce two separate scores, which are then combined by the ensemble component before they are processed by the threshold component to determine the outliers. The third case is the proposed NA where the dataset is input into an outlier detector, after which the scores are averaged before they were processed by the threshold component. The last case is a combination of the multiple-detector ensemble + NA, where two outlier detectors produce scores which are first combined by the ensemble and then post-processed by NA.

To summarize this paper's contribution, we use *k*-NN to post-process the existing outlier scores to produce more reliable and consistent scores. While there are already many *k*-NN-based methods, they all operate in the feature space. In contrast, NA operates in the score space by modifying existing scores without any additional information besides the neighborhood graph. The method is not limited to geographical data [34] or any other single application, but it can be applied in any application domain. It can improve any existing score-based outlier detectors or ensemble techniques, and it is not limited to use with $k$-NN-based outlier detectors.

We organize this paper as follows. In Section II, we recall several state-of-the-art outlier detectors from several categories. They later are used as our baseline detectors. In Section III, we introduce the proposed hypothesis and NA. The experimental setup is described in Section IV, and the results are shown in Section V. In Section VI, we describe our conclusions.

## II. OUTLIER DETECTORS

By constructing the *reference set* [1] for the calculation of outlier scores, outlier detectors can be grouped into global detectors and local detectors. Global detectors use all objects and local detectors use only a small subset of objects, such as $k$-NN, in the dataset as a reference set. We next review 12 well-known and state-of-the-art outlier detectors including six $k$-NN-based detectors.

In distance-based outlier detectors [12, 13, 14], outlier objects essentially should be located far away from other objects. The detector in [12] computes the distance between an object and its $k^{th}$ nearest neighbor as the outlier score. This detector is referred to as KNN [12]. A variant that evaluates the average distance to its all $k$ neighbors was proposed in [13]. The method in [15] calculates the distance to the average of its $k$-NN. It uses spatial features to determine the neighbors and the other features for the outlier detection.

Instead of considering distance, the detector in [14] counts the number of objects within a predefined distance threshold to the object. The count is used as the outlier score. *Outlier detection using indegree of nodes* (ODIN) in [13] is also based on the $k$-NN graph. It uses the time spent as another object's neighbor as the outlier score.

*Reverse unreachability* (NC as defined in [16]) is a detector based on representation. A given object is represented by $k$-NN with a weight matrix corresponding to the contribution from each neighbor. The negative weights carry information on the possibility of being outliers. The occurrence of negative weights is used as the outlier score.

*Mean-shift outlier detection* (MOD) [7, 17, 18] replaces an object with its $k$-NN's mean. This process is repeated three times. The distance between the original and the modified value of an object is the outlier score. This approach works well, especially when a dataset contains a large number of outliers [7].

In density-based detectors [19, 20], outlier objects have considerably lower densities than their neighbors. *Local outlier factor* (LOF) [16] evaluates the density of an object relative to that of its $k$-NN as the outlier score. In [21], it was reported to be the best-known detector when compared to the other 12 $k$-NN-based detectors.

The *minimum covariance determinant* (MCD) [22] is based on statistical analysis and is a robust estimator for evaluating the mean and covariance matrix. It finds 50% of objects with a covariance matrix having the smallest determinant. It then uses the difference from an object to the center of the objects as the outlier score.

*Isolation-based anomaly detection* (IFOREST as defined in [23]) builds trees over the dataset. It recursively separates the objects into two parts with a random threshold given a randomly selected feature. To remove the bias of randomness, it repeats the process several times. The average number of splits to isolate an object from other objects is its outlier score. An improved version of IFOREST can be found in [24].

*Support vector machine* (SVM) has been widely applied to pattern recognition tasks. *One class support vector machine* (OCSVM) [25] treats the objects as training data and creates a one-class model. The distance to the trained model is then used as the outlier score.

*Principal component analysis* (PCA) is an established data-mining technique. PCA can extract the principal structure of the data. The *principal-component-analysis-based outlier detection method* (PCAD) [26] reconstructs objects using the eigenvectors with reconstruction errors. The normalized errors are outlier scores.

*Angle-based outlier detection* (ABOD) [27] calculates the angles between objects. The variance of these angles is used as the outlier score. It was viewed as overcoming dimensionality better than distance-based measures in [27].

*Multiple-objective generative-adversarial active learning* (MO-GAAL) [28] is proposed to overcome the sparsity of data in high-dimensional space by generating additional data objects. MO-GAAL first trains a neural network to classify the generative and real-data objects. The outlier score is calculated as the possibility of the object being real.

*Copula-based outlier detector* (COPOD) [29, 30] predicts the tail probabilities of each object by constructing an empirical copula. The probability is used as the outlier score.

## III. NEIGHBORHOOD AVERAGING

In this section, we present the general framework of NA. In general, outlier detectors utilize different assumptions to produce outlier scores, such as distance or density. However, besides this assumption, we do not set any requirements but rely on the existing detectors and their assumptions.

### A. General averaging framework

The example in Fig. 1 shows that *similar objects should have similar outlier scores*. Although Detector 1 can find the two outliers (with the proper threshold), by plotting the outlier scores in Fig. 4, we can see there is a local peak in the distribution of the outlier scores, which does not match reality. Fig. 5 shows that the local peak will cause either a false positive or a false negative regardless of which threshold value is selected. It is therefore necessary to remove the local peak.

In a recommendation system [31], a related hypothesis for collaborative filtering techniques states that *similar users must/should have similar preferences.* Both of these hypotheses rely on defining the similarity of the objects in the feature space. However, there is one important difference between them. While collaborative filtering does not involve any score calculations, the definition of the outlier score is the key to outlier detection. Fig. 6 shows three types of similar objects.

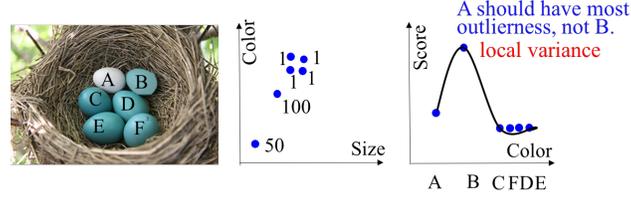

Fig. 4. Define local variance in outlier scores: relative outlier scores do not match the relative degree of being outliers.

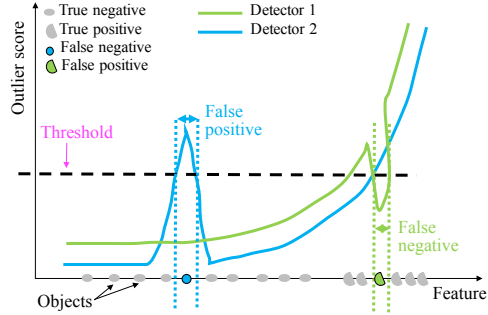

Fig. 5. Visualization of how the local variance (local peak) affects the accuracy of outlier detection. The blue line and green line have local variances. We can see that no matter how we adjust the threshold value, the local variance affects the accuracy by causing either a false positive or a false negative.

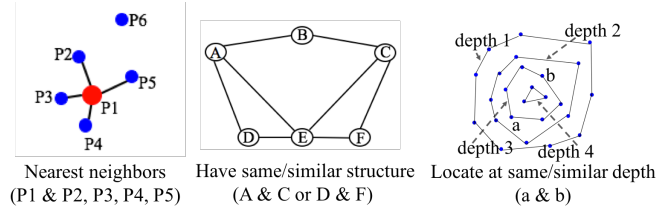

Fig. 6. Definition of the similarity of objects can be different with different data. In feature space, it can be based on the distance of objects (left); it can be based on the nodes' common neighbors (middle); and it can be based on which level the object is located in the structure (right).

### B. Neighborhood averaging (NA)

The proposed NA technique is simple: We take any baseline outlier detector and use it to compute the preliminary outlier score for every object. We then modify the objects' outlier scores in the neighborhood to be closer to one another to smooth the baseline outlier detectors' results. The main advantage of the technique is its applicability to any existing outlier detector or technique. While we use $k$-NN in this paper, it should be noted that any neighborhood model can also be applied.

Neighborhood averaging uses two steps to modify a given dataset X's outlier score S produced by any detector. In the first step, for each object $X_i$, NA looks for its $k$-NN: $k$-NN($X_i$). In the second step, NA revises each outlier score $S_i$, to be $S^*_i$, which is the average of the scores of its $k$-NN:

$$S^*_i \leftarrow \frac{1}{k+1}\left(S_i + \sum_{j=1}^{k} S_j\right); X_j \in k-\text{NN}(X_i), S_j \in S \qquad (1)$$

Algorithm 1 shows the pseudo-code and Fig. 7 demonstrates NA's two steps. Considering the red object (object B in Fig. 4), NA first searches its $k$-NN and then calculates the average scores of the neighbors. As a result, the peak in the outlier scores in Fig. 4 has been removed. The visualization examples with and without NA are shown in Fig. 8. We can see that the LOF detector (with $k = 40$) falsely detects many boundary objects as outliers (cross), but it succeeds after using NA.

Neighborhood averaging updates the outlier score of an object by the average of the scores of its neighbors. Where the object is also a neighbor of other objects, NA would be applied with multiple iterations and in each current iteration, only the score of the last iteration is used to revise each object's score.

---

**Algorithm 1:** NA(X, S, *k*)

---

**Input:** Dataset X, Outlier scores S, Neighborhood size *k*
**Output:** Revised outlier scores S*

FOR EACH X$_i$ ∈ X:

    *k*-NN(X$_i$) ← Find *k* nearest neighbors of X$_i$
    S*$_i$ ← average with scores of *k*-NN(X$_i$) according to (1)

---

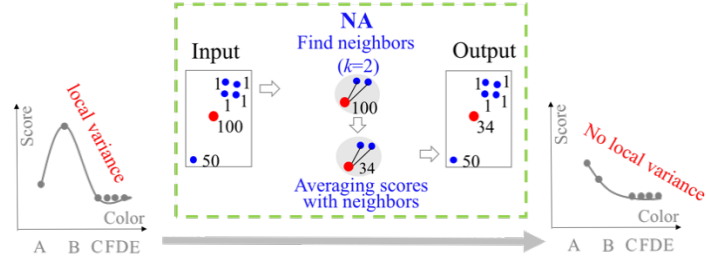

Fig. 7. Illustration of the averaging process: an object B from Fig. 4 (red), and all the outlier scores. NA first finds the 2 nearest objects of B and then calculates the average score within its neighbors as the revised score: (100+1+1)/3 = 34. As a result, the local peak has been successfully removed by NA.

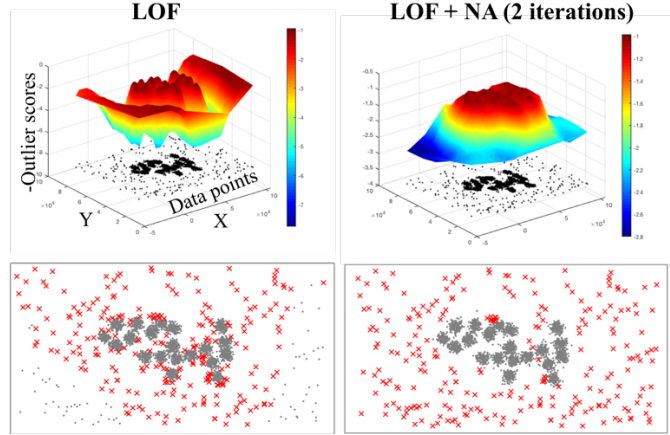

Fig. 8. Visualization of the outlier scores (top) and the detected outliers (bottom). The results at left and right are given by the LOF detector with and without NA, respectively (*k* = 40). LOF falsely detects boundary objects as outliers (cross) as evaluated on a noisy A1 dataset [33], while NA improves the result of the LOF significantly.

### C. Discussion

In this section, we discuss the proposed NA and closely related techniques. One is the *k*-NN classifier, which also looks for neighborhood objects when classifying objects. The difference is that the *k*-NN classifier is a supervised method, but NA is not.

Another related technique is the mean-shift technique in [7], which is also widely applied in image processing [32]. Neighborhood averaging can be repeated several times and the process iteratively replaces an object's outlier score with its neighbors' mean scores. This process is close to the mean-shift process [11]. The difference is that mean-shift modifies the feature values of the objects whereas NA modifies their scores.

All *k*-NN-based outlier detectors are related to one another in the sense that they use *k*-NN as their key component. However, how they use *k*-NN differs. In general, all *k*-NN-based detectors use *k*-NN to *produce* the outlier scores for the objects, as shown at the top of Fig. 9, while NA uses *k*-NN to *revise* the outlier scores produced by any detector, including all *k*-NN-based detectors, as shown at the bottom of Fig. 9. Using *k*-NN as a *detector* to produce outlier scores is a well-known approach but it is novel to use it as a *post-processing technique* for tuning the score.

Outlier scores
of detector d:  $S^d = f^d(\text{data})$

Revised outlier
Score of ensemble:  $R = g^d(S^1, \dots, S^d)$

Revised outlier
Score of NA:  $R^d = h^d(\text{data}, S^d)$

Fig. 9. Difference between *k*-NN-based detectors, ensembles, and NA

It is worth noting that other detectors in [7, 13, 15] also utilize $k$-NN and the *average* operation. However, these are stand-alone detectors and cannot be an add-on to detectors, while NA is an add-on to other detectors and cannot be used as a stand-alone detector.

Ensemble techniques are also related and have the *combination* operation. Besides this commonality, NA has three fundamental differences. First, ensemble techniques combine several poor detectors to obtain a better one [1], as shown in the revised outlier score in the ensemble in Fig. 9, while NA removes local variance. Second, ensemble techniques need to compute the outlier score for the same object multiple times, while NA does not. Third, ensemble techniques cannot be applied to a single detector, but NA can.

Neighborhood averaging and ensemble techniques are not exclusive, and they can be applied jointly. Their similarity is that both aim to smooth the outlier scores; ensemble operates across the detectors while NA operates across the objects. Considering the two detectors (the blue and green lines) in Fig. 5, ensemble techniques can improve these two poorly performing detectors only when the two peaks happen in the same location (objects) and with a proper difference.

It is worthwhile to note that NA may be suitable for other score-based data-mining tasks. This is because similar input should have similar output. If similar input does not provide similar output, the model is not consistent. If we define any techniques having the operation of *combination* as ensemble techniques, we already have feature ensemble (feature bagging), detector ensemble, parameter ensemble, and object ensemble (NA). These ensembles should be applicable to data-mining tasks other than outlier detection.

Recently, Ke et al. [34] proposed a method called group similarity system (GSS) for unsupervised outlier detection and Yang et al. [35] proposed a data pre-processing technique called neighborhood representative (NR) to detect collective outliers using exiting outlier detectors. GSS partitions the data into non-overlapped groups and judges the groups as outliers by considering the mean of the outlier scores of the objects in each group. NR scores the representative objects sampled from each group and judges the groups as outliers by considering the scores of the representative objects in each group. Neighborhood averaging is not used for collective outliers but for individual outliers, making it different from GSS or NR.

## IV. EXPERIMENTAL SETUP

We used nine public, real-world, semantically meaningful static datasets, which can be found in UCI repository datasets or [21]. The information in the datasets varies from 8 to 259. They contain outliers ranging from 0.40% to 75.40% and have objects ranging from 195 to 60,632 as summarized in Table I. For preprocessing, all data were scaled by subtracting the mean and dividing by standard deviation for each attribute.

Table I DATASET INFORMATION

| Name | Objects | Dim | Outlier | Outlier objects | Normality objects |
|---|---|---|---|---|---|
| KDD-Cup99 | 60,632 | 38 | 246 | Intrusion connections | Normal connections |
| Stamps | 340 | 9 | 340 | Forged stamps | Genuine stamps |
| PageBlocks | 5,393 | 10 | 510 | Text block | Other types of block |
| Cardiotocography | 2,114 | 21 | 466 | Suspect or pathological people | Healthy people |
| Pima | 768 | 8 | 268 | People with diabetes | Healthy people |
| SpamBase | 4,207 | 57 | 1679 | Spam emails | Non-spam emails |
| HeartDisease | 270 | 13 | 120 | People with heart problem | Healthy people |
| Arrhythmia | 450 | 259 | 206 | Patients with arrhythmia | Healthy people |
| Parkinson | 195 | 22 | 147 | Patients with Parkinson's disease | Healthy people |

The outlier detectors' performance was measured mainly by *area under* the *receiver operating characteristic* (ROC) *curve* (AUC). The ROC curve was drawn by plotting the true positive rate against the false positive rate over various threshold values. The AUC was a single value ranging from 0 to 1. The bigger the value was, the better the performance.

While AUC measured the average performance, we also tested the performance when a selected thresholding method was applied. For the threshold component, we used the known number of outliers in the dataset. This is known as the *top-k method*. The result was measured by the *F1-score*, which was the average of *precision* and *recall*. Precision is the ability to minimize false positives and recall is the ability to find all the positive samples.

For $k$-NN-based outlier detectors, we used the value of $k$, which provided the best results when $k$ ranged from 2 to 100. The default parameters found in the literature are used for the other detectors.

The proposed NA was tested with all values of $k$ from 1 to 100. We used $k = 100$ as the default value. Neighborhood averaging was iterated 10 times to study the effect of *iterations*.

## V. RESULTS

### A. The overall effect of NA

We varied the neighborhood size $k$ in NA from 1 to 100 to find the best results and compared them with the results obtained using the default value $k = 100$. The average AUC and F1-score results are summarized in Table II. The AUC results per dataset are summarized in Table III. Based on the results, we can make the following observations.

First, based on the AUC results in Table II, the proposed NA significantly improved all the detection results. On average, all the detectors evaluated for all the datasets improved by +0.04 (from 0.70 to 0.74) with the default $k$, and +0.06 with the best $k$. We can make a similar observation about AUC for the F1-score. Neighborhood averaging improved all outlier detectors by +0.02 (from 0.73 to 0.75) on average when using the default value of $k$, and by +0.03 when using the best value of $k$.

Second, NA provided the most AUC improvement with the NC detector, from 0.62 to 0.77. The most significant individual improvement was +0.28 for **HeartDisease** and **KDD-Cup99**. This observation is interesting, as NC was originally one of the worst detectors. However, when used with NA, it became competitive. This indicates that NC and NA utilize different properties and are complementary. It also suggests that the poorly performing detectors evaluated previously may have been seriously underestimated.

Third, the default setting with $k = 100$ performed almost as well as the best $k$. This shows that NA is robust on the choice of the parameter $k$.

Fourth, as shown in the rows of data labeled *original* in Table I, except MO_GAAL, without using NA the average AUC of detectors has a range from 0.62 to 0.75. However, with NA, the range becomes much smaller, from 0.75 to 0.79. This indicates that when NA was not used the choice of detector mattered, but when NA was used it mattered less. This may pose a challenge to the generally accepted idea that *the data model is the most important factor* [1]. For MO_GAAL, the ROC AUC is near 0.50, which is close to random guesses. This may be because MO_GAAL needed more samples to train the neural network.

In Table III, we can see that all detectors for all datasets improved for both the default $k$ and the best $k$. The only exception is the result for **Arrhythmia**, which weakened by -0.02 when using default $k$. Most datasets improved from +0.03 to +0.15 on average. The most significant individual improvement was for **HeartDisease**, which was +0.17 on average. Neighborhood averaging did not help much with the datasets containing only a few outliers or when the original detector already performed well. For example, MOD, KNN, IFOREST, OCVSCM, and PCAD all achieved AUC = 0.99 for **KDD-Cup99**.

### B. Effect of the iterations

Neighborhood averaging can be iterated several times. Next, we varied the *iteration* parameter from 1 to 10 times to study its effect on the result. The value *iteration* = 0 corresponds to the original detector without NA. The average AUC results of all detectors evaluated for all datasets, a selected detector (MOD), and a selected dataset (**HeartDisease**) are summarized in Table IV, Table V, and Table VI, respectively.

The average results in Table IV show the first iteration achieved the most improvement (+0.06). The second iteration achieved further improvement (+0.01) but beyond that, the effect remained rather small (<+0.03). However, by applying NA for multiple iterations the performance was improved from 0.70 to 0.79 UAC.

The results for the individual datasets with MOD are reported in Table V. All the datasets evaluated with the MOD detector were improved except **Arrhythmia**, which started to deteriorate during the second iteration. This might have been caused by the so-called *curse of dimensionality* in high-dimensional data, as **Arrhythmia** has 259 dimensions, while all the other datasets had 60 or fewer. Most other datasets were improved even when they were iterated 10 times. Another exception was **Pima**, for which the result started to deteriorate after the fourth iteration. This indicated that the *iteration* parameter needed to be tuned according to the datasets if an optimal value was desired. To be conservative, we set the default value as *iteration* = 1 despite knowing that some datasets, such as **SpamBase** and **HeartDisease,** would benefit from more iterations.

The results for the individual detectors with **HeartDisease** are reported in Table VI. It shows all detectors can benefit from *iteration* = 2.

To summarize, we conclude that the optimal number of iterations in applying NA depends on the dataset and the detector used and it is not trivial to optimize. Our recommended choice is *iteration* = 1.

### C. The value of k

To study the effect of $k$ in NA, we varied it from 1 to 100. The average AUC values across all the datasets are shown in Fig. 10. The results on a selected individual dataset (**HeartDisease**) are also shown in Fig. 11. The value $k = 1$ corresponds to the original detector without NA.

The results show that when increasing $k$, all detectors improved and reached their best performance with $k = 100$. We therefore recommend $k = 100$ as the default value.

Neighborhood averaging is proposed as an independent component to improve single outlier detectors. We notice that all $k$-NN-based outlier detectors also need to select the value of $k$. We considered using the same $k$ value both for the baseline detectors and for NA. We performed additional experiments with the $k$-NN-based detectors. We varied $k$ from 3 to 100 to find the best AUC.

The average results over all datasets are summarized in Table VII. They show that NA significantly improved the detectors by +0.05 on average. Most improvement is achieved with NC (+0.11). Further minimal improvements might be achieved with some datasets if $k$ was increased further. However, some datasets do not have enough data to go much beyond 100, and the results would eventually start to degrade. The main result was that we can achieve good performance with rather small $k$ values.

TABLE II AVERAGE AUC AND F1-SCORE FOR ALL DATASETS

| Name | AUC | | | F1-score | | |
|---|---|---|---|---|---|---|
| | Original | NA (k) | | Original | NA (k) | |
| | | default | best | | default | best |
| MOD[7] | 0.73 | 0.77 | 0.78 | 0.74 | 0.76 | 0.77 |
| LOF[19] | 0.71 | 0.76 | 0.77 | 0.74 | 0.76 | 0.78 |
| ODIN[13] | 0.67 | 0.74 | 0.75 | 0.71 | 0.75 | 0.76 |
| NC[16] | 0.62 | 0.74 | 0.77 | 0.71 | 0.75 | 0.77 |
| KNN[12] | 0.75 | 0.76 | 0.79 | 0.74 | 0.76 | 0.79 |
| ABOD[27] | 0.69 | 0.72 | 0.75 | 0.73 | 0.75 | 0.75 |
| MCD[22] | 0.71 | 0.75 | 0.77 | 0.72 | 0.75 | 0.76 |
| IFOREST[23] | 0.74 | 0.77 | 0.79 | 0.74 | 0.76 | 0.77 |
| OCSVM[25] | 0.71 | 0.75 | 0.76 | 0.71 | 0.75 | 0.76 |
| PCAD[26] | 0.72 | 0.75 | 0.76 | 0.73 | 0.75 | 0.75 |
| MO_GAAL[28] | 0.55 | 0.58 | 0.60 | 0.67 | 0.69 | 0.69 |
| COPOD[30] | 0.75 | 0.77 | 0.79 | 0.76 | 0.77 | 0.78 |
| **AVG** | **0.70** | **0.74** | **0.76** | **0.73** | **0.75** | **0.76** |

TABLE IV AVERAGE AUC RESULTS FOR ALL DATASETS

| Detector | 0 | 1 | 2 | 3 | 4 | 5 | 6 | 7 | 8 | 9 | 10 |
|---|---|---|---|---|---|---|---|---|---|---|---|
| | | | | | | Iteration | | | | | |
| MOD | 0.73 | 0.78 | 0.80 | 0.81 | 0.81 | 0.81 | 0.81 | 0.81 | 0.81 | 0.81 | 0.81 |
| LOF | 0.71 | 0.77 | 0.78 | 0.79 | 0.80 | 0.80 | 0.80 | 0.80 | 0.80 | 0.80 | 0.80 |
| ODIN | 0.67 | 0.75 | 0.77 | 0.78 | 0.79 | 0.79 | 0.80 | 0.80 | 0.80 | 0.80 | 0.80 |
| NC | 0.62 | 0.77 | 0.76 | 0.76 | 0.75 | 0.75 | 0.74 | 0.74 | 0.74 | 0.74 | 0.74 |
| KNN | 0.75 | 0.79 | 0.81 | 0.81 | 0.81 | 0.81 | 0.81 | 0.81 | 0.81 | 0.81 | 0.81 |
| ABOD | 0.69 | 0.75 | 0.78 | 0.79 | 0.79 | 0.79 | 0.79 | 0.79 | 0.79 | 0.79 | 0.79 |
| MCD | 0.71 | 0.77 | 0.78 | 0.79 | 0.79 | 0.79 | 0.79 | 0.79 | 0.79 | 0.79 | 0.79 |
| IFOREST | 0.74 | 0.79 | 0.80 | 0.82 | 0.82 | 0.82 | 0.82 | 0.82 | 0.82 | 0.82 | 0.83 |
| OCSVM | 0.71 | 0.76 | 0.79 | 0.81 | 0.82 | 0.81 | 0.82 | 0.82 | 0.82 | 0.82 | 0.82 |
| PCAD | 0.72 | 0.76 | 0.77 | 0.78 | 0.78 | 0.78 | 0.78 | 0.78 | 0.78 | 0.78 | 0.78 |
| MO_GAAL | 0.55 | 0.60 | 0.61 | 0.61 | 0.61 | 0.61 | 0.61 | 0.61 | 0.61 | 0.61 | 0.62 |
| COPOD | 0.76 | 0.79 | 0.82 | 0.83 | 0.84 | 0.84 | 0.84 | 0.85 | 0.85 | 0.85 | 0.85 |
| **AVG** | 0.70 | 0.76 | 0.77 | 0.78 | 0.78 | 0.78 | 0.78 | **0.79** | **0.79** | **0.79** | **0.79** |

TABLE III AUC IMPROVEMENT PER OUTLIER DETECTOR PER DATASET

| Dataset outliers | KDD-Cup99 0.4% | Stamps 9.1% | PageBlocks 10.2% | Cardio. 22.2% | Pima 34.9% | SpamBase 39.4% | HeartDisease 44.4% | Arrhythmia 45.8% | Parkinson 75.4% | AVG | DIFF |
|---|---|---|---|---|---|---|---|---|---|---|---|
| **Original** | | | | | | | | | | | |
| MOD | **0.99** | 0.90 | 0.91 | 0.54 | 0.68 | 0.55 | 0.62 | 0.74 | 0.64 | 0.73 | - |
| LOF | 0.84 | 0.89 | 0.91 | 0.59 | 0.69 | 0.49 | 0.67 | 0.73 | 0.60 | 0.71 | - |
| ODIN | 0.81 | 0.83 | 0.79 | 0.61 | 0.63 | 0.52 | 0.61 | 0.72 | 0.53 | 0.67 | - |
| NC | 0.69 | 0.68 | 0.70 | 0.57 | 0.57 | 0.55 | 0.58 | 0.67 | 0.56 | 0.62 | - |
| KNN | **0.99** | 0.91 | **0.92** | 0.55 | **0.73** | 0.57 | 0.68 | 0.74 | **0.66** | 0.75 | - |
| ABOD | 0.86 | 0.87 | 0.85 | 0.48 | 0.70 | 0.42 | 0.65 | 0.73 | 0.64 | 0.69 | - |
| MCD | 0.97 | 0.85 | **0.92** | 0.49 | 0.68 | 0.46 | 0.64 | 0.72 | 0.64 | 0.71 | - |
| IFOREST | **0.99** | 0.86 | 0.90 | 0.70 | 0.67 | 0.64 | 0.65 | **0.76** | 0.47 | 0.74 | - |
| OCSVM | **0.99** | 0.87 | 0.91 | 0.70 | 0.62 | 0.53 | 0.58 | 0.74 | 0.43 | 0.71 | - |
| PCAD | **0.99** | 0.90 | 0.90 | **0.75** | 0.63 | 0.55 | 0.62 | 0.73 | 0.38 | 0.72 | - |
| MO_GAAL | 0.55 | 0.63 | 0.56 | 0.56 | 0.50 | **0.73** | 0.41 | 0.50 | 0.50 | 0.55 | - |
| COPOD | **0.99** | **0.93** | 0.88 | 0.66 | 0.65 | 0.69 | **0.69** | **0.76** | 0.54 | 0.75 | - |
| **AVG** | 0.89 | 0.84 | 0.85 | 0.60 | 0.65 | 0.56 | 0.62 | 0.71 | 0.55 | 0.70 | - |
| **NA (default)** | | | | | | | | | | | |
| MOD | **0.99** | **0.93** | 0.91 | 0.52 | 0.76 | 0.59 | 0.77 | 0.72 | 0.72 | 0.77 | 0.04 |
| LOF | 0.88 | **0.93** | **0.94** | 0.58 | 0.76 | 0.67 | 0.78 | **0.73** | 0.55 | 0.74 | 0.05 |
| ODIN | 0.83 | **0.93** | 0.83 | 0.74 | 0.74 | 0.57 | 0.75 | 0.70 | 0.56 | 0.74 | 0.07 |
| NC | 0.97 | 0.92 | 0.86 | 0.80 | 0.61 | 0.61 | **0.85** | 0.70 | 0.32 | 0.74 | 0.12 |
| KNN | 0.88 | 0.92 | 0.90 | 0.52 | **0.77** | 0.61 | 0.82 | 0.72 | 0.73 | 0.76 | 0.01 |
| ABOD | 0.85 | 0.90 | 0.79 | 0.41 | 0.78 | 0.36 | 0.85 | 0.70 | **0.80** | 0.72 | 0.03 |
| MCD | **0.99** | 0.92 | 0.93 | 0.47 | 0.74 | 0.43 | 0.81 | 0.70 | 0.76 | 0.75 | 0.04 |
| IFOREST | 0.98 | 0.92 | 0.87 | 0.74 | 0.73 | 0.65 | 0.81 | 0.70 | 0.54 | 0.77 | 0.03 |
| OCSVM | 0.98 | 0.92 | 0.90 | 0.75 | 0.68 | 0.56 | 0.72 | 0.72 | 0.56 | 0.75 | 0.05 |
| PCAD | 0.98 | 0.92 | 0.89 | **0.81** | 0.69 | 0.58 | 0.76 | 0.70 | 0.40 | 0.75 | 0.03 |
| MO_GAAL | 0.58 | 0.80 | 0.42 | 0.50 | 0.52 | **0.74** | 0.63 | 0.50 | 0.58 | 0.58 | 0.03 |
| COPOD | 0.95 | **0.93** | 0.84 | 0.68 | 0.70 | 0.69 | 0.82 | 0.71 | 0.60 | 0.77 | 0.01 |
| **AVG** | 0.90 | 0.91 | 0.84 | 0.63 | 0.71 | 0.59 | 0.78 | 0.69 | 0.59 | 0.74 | 0.04 |
| **NA (best)** | | | | | | | | | | | |
| MOD | **0.99** | **0.95** | 0.92 | 0.55 | 0.76 | 0.59 | 0.77 | 0.74 | 0.74 | 0.78 | 0.05 |
| LOF | 0.88 | 0.94 | **0.94** | 0.59 | 0.76 | 0.67 | 0.78 | 0.74 | 0.60 | 0.77 | 0.06 |
| ODIN | 0.83 | 0.94 | 0.83 | 0.74 | 0.74 | 0.57 | 0.75 | 0.72 | 0.58 | 0.74 | 0.07 |
| NC | 0.97 | 0.92 | 0.86 | 0.80 | 0.61 | 0.62 | **0.86** | 0.72 | 0.58 | 0.77 | 0.15 |
| KNN | **0.99** | **0.95** | 0.92 | 0.55 | 0.77 | 0.61 | 0.82 | 0.74 | 0.75 | 0.79 | 0.04 |
| ABOD | 0.93 | 0.94 | 0.85 | 0.48 | **0.78** | 0.39 | **0.86** | 0.73 | 0.80 | 0.75 | 0.06 |
| MCD | **0.99** | 0.92 | 0.93 | 0.49 | 0.74 | 0.48 | 0.81 | 0.73 | **0.82** | 0.77 | 0.06 |
| IFOREST | **0.99** | 0.94 | 0.90 | 0.74 | 0.73 | 0.65 | 0.81 | **0.76** | 0.54 | 0.78 | 0.05 |
| OCSVM | **0.99** | 0.93 | 0.91 | 0.75 | 0.68 | 0.56 | 0.73 | 0.74 | 0.56 | 0.76 | 0.05 |
| PCAD | **0.99** | 0.94 | 0.90 | **0.81** | 0.69 | 0.58 | 0.77 | 0.73 | 0.40 | 0.76 | 0.04 |
| MO_GAAL | 0.60 | 0.80 | 0.56 | 0.50 | 0.57 | **0.74** | 0.63 | 0.50 | 0.60 | 0.60 | 0.05 |
| COPOD | **0.99** | **0.95** | 0.88 | 0.69 | 0.70 | 0.71 | 0.82 | **0.76** | 0.60 | 0.79 | 0.03 |
| **AVG** | 0.93 | 0.93 | 0.87 | 0.64 | 0.71 | 0.60 | 0.78 | 0.72 | 0.62 | 0.76 | 0.06 |

TABLE V AUC RESULTS OF MOD DETECTOR

| Dataset | 0 | Iteration | | | | | | | | | |
|---|---|---|---|---|---|---|---|---|---|---|---|
| | | 1 | 2 | 3 | 4 | 5 | 6 | 7 | 8 | 9 | 10 |
| KDD-Cup99 | 0.99 | 0.99 | 0.99 | 0.99 | 0.99 | 0.99 | 0.99 | 0.99 | 0.99 | 0.99 | 0.99 |
| Stamps | 0.90 | 0.95 | 0.95 | 0.95 | 0.96 | 0.96 | 0.96 | 0.95 | 0.95 | 0.95 | 0.95 |
| PageBlocks | 0.91 | 0.92 | 0.92 | 0.92 | 0.92 | 0.92 | 0.92 | 0.92 | 0.92 | 0.92 | 0.92 |
| Cardio. | 0.54 | 0.55 | 0.55 | 0.55 | 0.55 | 0.55 | 0.55 | 0.55 | 0.55 | 0.56 | 0.56 |
| Pima | 0.68 | 0.76 | 0.78 | 0.78 | 0.77 | 0.76 | 0.76 | 0.75 | 0.75 | 0.75 | 0.75 |
| SpamBase | 0.55 | 0.59 | 0.65 | 0.69 | 0.72 | 0.74 | 0.75 | 0.75 | 0.76 | 0.76 | 0.77 |
| HeartDisease | 0.62 | 0.77 | 0.86 | 0.89 | 0.90 | 0.90 | 0.90 | 0.91 | 0.91 | 0.91 | 0.91 |
| Arrhythmia | 0.74 | 0.74 | 0.71 | 0.69 | 0.67 | 0.65 | 0.65 | 0.64 | 0.64 | 0.64 | 0.64 |
| Parkinson | 0.64 | 0.74 | 0.80 | 0.83 | 0.84 | 0.84 | 0.85 | 0.85 | 0.85 | 0.85 | 0.85 |
| **AVG** | **0.73** | **0.78** | **0.80** | **0.81** | **0.81** | **0.81** | **0.81** | **0.81** | **0.81** | **0.81** | **0.81** |

TABLE VI AUC RESULTS ON **HEARTDISEASE** DATASET

| Detector | 0 | Iteration | | | | | | | | | |
|---|---|---|---|---|---|---|---|---|---|---|---|
| | | 1 | 2 | 3 | 4 | 5 | 6 | 7 | 8 | 9 | 10 |
| MOD | 0.62 | 0.77 | 0.86 | 0.89 | 0.90 | 0.90 | 0.90 | 0.91 | 0.91 | 0.91 | 0.91 |
| LOF | 0.67 | 0.78 | 0.85 | 0.89 | 0.90 | 0.90 | 0.90 | 0.91 | 0.91 | 0.91 | 0.91 |
| ODIN | 0.61 | 0.75 | 0.85 | 0.89 | 0.90 | 0.90 | 0.90 | 0.91 | 0.91 | 0.91 | 0.91 |
| NC | 0.58 | 0.86 | 0.90 | 0.91 | 0.91 | 0.91 | 0.91 | 0.91 | 0.91 | 0.91 | 0.91 |
| KNN | 0.82 | 0.87 | 0.89 | 0.90 | 0.90 | 0.90 | 0.91 | 0.91 | 0.91 | 0.91 | 0.91 |
| ABOD | 0.65 | 0.86 | 0.89 | 0.90 | 0.90 | 0.90 | 0.91 | 0.91 | 0.91 | 0.91 | 0.91 |
| MCD | 0.61 | 0.81 | 0.87 | 0.89 | 0.90 | 0.90 | 0.90 | 0.91 | 0.91 | 0.91 | 0.91 |
| IFOREST | 0.65 | 0.81 | 0.87 | 0.89 | 0.90 | 0.90 | 0.91 | 0.91 | 0.91 | 0.91 | 0.91 |
| OCSVM | 0.58 | 0.73 | 0.84 | 0.89 | 0.90 | 0.90 | 0.90 | 0.91 | 0.91 | 0.91 | 0.91 |
| PCAD | 0.62 | 0.77 | 0.86 | 0.89 | 0.90 | 0.90 | 0.90 | 0.91 | 0.91 | 0.91 | 0.91 |
| MO_GAAL | 0.41 | 0.63 | 0.64 | 0.63 | 0.62 | 0.62 | 0.61 | 0.61 | 0.61 | 0.61 | 0.61 |
| COPOD | 0.69 | 0.82 | 0.88 | 0.90 | 0.90 | 0.91 | 0.91 | 0.91 | 0.91 | 0.91 | 0.91 |
| **AVG** | **0.62** | **0.78** | **0.85** | **0.87** | **0.88** | **0.88** | **0.88** | **0.88** | **0.89** | **0.88** | **0.89** |

TABLE VIII AUC DIFFERENCE RESULTS FOR ALL DATASETS FOR AVERAGE JOINTLY WORKING WITH NA

| Detector combination | Post-processing | KDD-Cup99 0.4% | Stamps 9.1% | PageBlocks 10.2% | Cardio. 22.2% | Pima 34.9% | SpamBase 39.4% | HeartDisease 44.4% | Arrhythmia 45.8% | Parkinson 75.4% | AVG |
|---|---|---|---|---|---|---|---|---|---|---|---|
| MOD and KNN | *Average* | **0.99** | 0.91 | **0.92** | 0.55 | 0.73 | 0.57 | 0.68 | **0.74** | 0.66 | 0.75 |
| | +NA (default $k$) | **0.99** | 0.92 | 0.90 | 0.52 | **0.77** | 0.61 | 0.82 | 0.72 | 0.73 | 0.77 |
| | +NA (best $k$) | **0.99** | **0.95** | **0.92** | 0.55 | **0.77** | 0.61 | 0.82 | **0.74** | **0.75** | **0.79** |
| ODIN and NC | *Average* | 0.81 | 0.83 | 0.79 | 0.61 | 0.63 | 0.52 | 0.61 | 0.72 | 0.53 | 0.67 |
| | +NA (default $k$) | 0.81 | 0.93 | 0.83 | 0.74 | 0.74 | 0.57 | 0.75 | 0.70 | 0.56 | 0.74 |
| | +NA (best $k$) | 0.83 | 0.94 | 0.83 | 0.74 | 0.74 | 0.57 | 0.75 | 0.72 | 0.58 | 0.75 |
| MOD and NC | *Average* | 0.69 | 0.68 | 0.70 | 0.57 | 0.57 | 0.55 | 0.58 | 0.67 | 0.56 | 0.62 |
| | +NA (default $k$) | 0.87 | 0.92 | 0.86 | **0.80** | 0.61 | 0.61 | 0.85 | 0.70 | 0.32 | 0.73 |
| | +NA (best $k$) | 0.87 | 0.92 | 0.86 | **0.80** | 0.61 | **0.62** | **0.86** | 0.72 | 0.58 | 0.76 |
| MOD, ODIN, NC, and KNN | *Average* | 0.99 | 0.90 | 0.91 | 0.54 | 0.68 | 0.55 | 0.62 | **0.74** | 0.64 | 0.73 |
| | +NA (default $k$) | **0.99** | 0.93 | 0.91 | 0.52 | 0.76 | 0.59 | 0.77 | 0.72 | 0.72 | 0.77 |
| | +NA (best $k$) | **0.99** | **0.95** | **0.92** | 0.55 | 0.76 | 0.59 | 0.77 | **0.74** | 0.74 | 0.78 |
| All twelve detectors | *Average* | 0.89 | 0.84 | 0.82 | 0.60 | 0.65 | 0.56 | 0.62 | 0.71 | 0.55 | 0.69 |
| | +NA (default $k$) | 0.91 | 0.91 | 0.82 | 0.63 | 0.71 | 0.59 | 0.77 | 0.69 | 0.59 | 0.73 |
| | +NA (best $k$) | 0.93 | 0.93 | 0.84 | 0.64 | 0.71 | 0.60 | 0.77 | 0.72 | 0.62 | 0.75 |

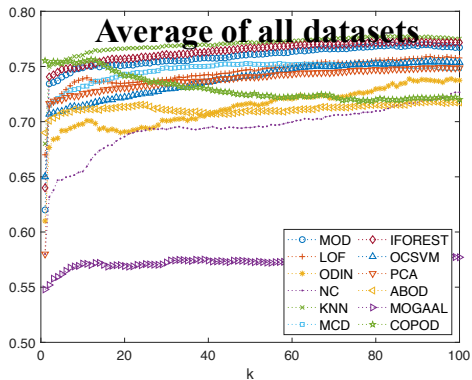

Fig. 10. Average AUC results for all datasets with varying $k$.

TABLE VII AVERAGE AUC RESULTS FOR ALL DATASETS WITH $K$ IN NA EQUALING TO $k$ IN $K$-NN BASED DETECTORS

| Detector | Original | | NA | |
|---|---|---|---|---|
| | Default $k$ | Best $k$ | Default $k$ | Best $k$ |
| MOD | 0.71 | 0.73 | **0.75** | **0.77** |
| LOF | 0.70 | 0.71 | 0.74 | 0.75 |
| ODIN | 0.66 | 0.67 | 0.73 | 0.74 |
| NC | 0.60 | 0.62 | 0.66 | 0.73 |
| KNN | **0.72** | **0.75** | **0.75** | **0.77** |
| ABOD | 0.68 | 0.69 | 0.71 | 0.74 |
| **AVG** | 0.68 | 0.70 | 0.72 | 0.75 |

TABLE IX AVERAGE AUC IMPROVEMENT AND EXTRA COMPUTING TIME USING NA FOR ALL DATASETS

| Detectors | | AUC improvement | Extra computing time |
|---|---|---|---|
| Category | Name | | |
| $k$-NN based | MOD | 7% | 2% |
| | LOF | 8% | 5% |
| | ODIN | 11% | 4% |
| | NC | 24% | 2% |
| | KNN | 5% | 5% |
| | ABOD | 11% | 5% |
| | **AVG** | **11%** | **4%** |
| Other - | MCD | 8% | 408% |
| | IFOREST | 6% | 891% |
| | OCSVM | 8% | 32% |
| | PCAD | 6% | 9800% |
| | MO_GAAL | 6% | >1% |
| | COPOD | 5% | 1583% |
| | **AVG** | **7%** | **2543%** |

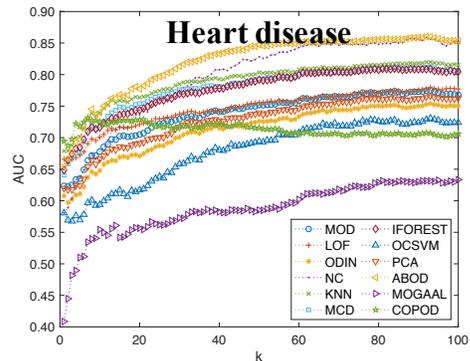

Fig. 11. AUC results on **HeartDisease** with varying $k$.

### D. Outlier ensembles

Next, we tested the effect of augmentation on NA with an existing outlier ensemble technique. We used the average ensemble [1] method, with different baseline detector combinations. Results are summarized in Table VIII.

We can observe that the results of the outlier ensemble depend on the quality of the individual detectors. The best results are obtained by the combination of MOD and KNN, which reaches 0.75. Combining all 12 detectors would reach only 0.69.

When applying NA jointly with the outlier ensemble, we observed the following. First, no matter which combination was used, NA always improved the result of the ensemble. Second, the best combination no longer depended on the quality of the individual detector. The best combination (MOD and KNN) is based on one of the weaker baseline detectors among those tested. This combination with NA reached the overall best result of 0.79, which is very close to the result (0.77) reached without optimizing the parameter $k$. This indicates that NA provides a strong complementary component to ensemble.

### E. Complementary to NR

As our previous work NR was a data preprocessing method to improve detectors, we wanted to know if NA as outlier score post-processing method could further improve NR. We tested LOF, NR+LOF, LOF+NA, and NR+LOF+NA by setting their parameter $k$ to be the same value and the results were plotted in Fig. 12 with **Parkinson**, **HeartDiease**, and **Spambase** datasets. From Fig. 12, we could observe a relative larger k, closed to 200, was good when NR and NA were jointly used. NR+NA+LOF could further improve NR 31% relative to NR+LOF on average (0.88 vs. 0.71 AUC) as shown on the right of Fig. 12. It was noteworthy that the performance of LOF on **Spambase** dataset with 0.49 AUC was close to random guessing, but when jointly used with NR and NA, it could even achieve 0.81 AUC. Another important observation was that the performance of **Parkinson** and **HeartDiease** of NR+LOF+NA could be more than 0.90 AUC compared to the LOF's results of less than 0.70, which was far better than any unsupervised results of existing literature, to our best knowledge. In a word, NA is complementary to NR significantly.

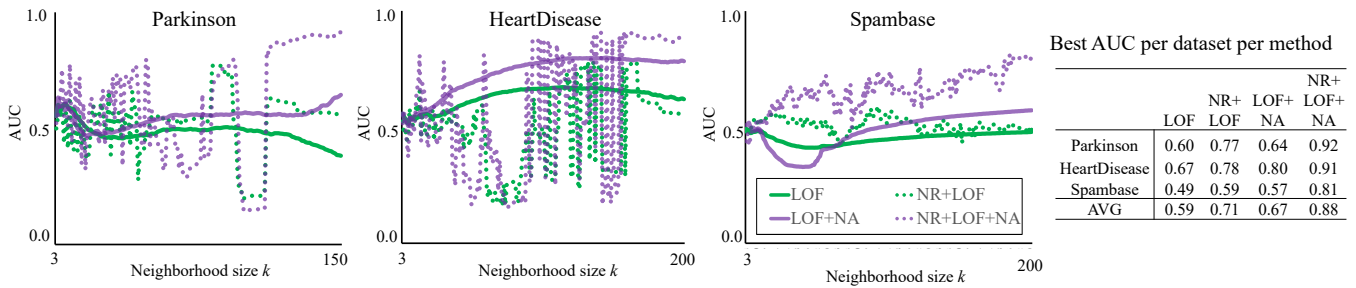

Fig. 12. Experiment results of LOF, NR+LOF, LOF+NA, and NR+LOF+NA ranging $k$. NA is complementary to NR.

| | LOF | NR+LOF | LOF+NA | NR+LOF+NA |
|---|---|---|---|---|
| Parkinson | 0.60 | 0.77 | 0.64 | 0.92 |
| HeartDiease | 0.67 | 0.78 | 0.80 | 0.91 |
| Spambase | 0.49 | 0.59 | 0.57 | 0.81 |
| AVG | 0.59 | 0.71 | 0.67 | 0.88 |

Best AUC per dataset per method

### F. Computational complexity

Neighborhood averaging requires O($NlogN$) calculations using KD-tree in low dimensions (D<20) and Ball-tree in higher dimensions (D>20) to find $k$-NN. However, since NA serves as a post-processing step, we care more about its gain relative to its additional cost. Table IX shows the average extra computing time and the average AUC improvement over all datasets.

Table IX shows that the $k$-NN-based detectors need only 4% extra time but can improve by 11% in AUC on average. Non-$k$-NN-based detectors are usually significantly faster and need 2,543% extra time to reach an average improvement of 7% in AUC. The main reason is that the $k$-NN-based detectors have already calculated the $k$-NN, which NA can directly utilize.

### G. Discussion and limitations

**Discussion.** Neighborhood averaging is not meant to be a stand-alone detector; rather, it is an add-on to any existing score-based outlier detector used to enhance its performance as shown in the neighborhood attention example in Fig. 3. The add-on does not increase the complexity of $k$-NN-based detectors as shown in section V-E, but it can bring significant improvement as shown in section V-A. Neighborhood averaging has only one parameter $k$ to tune, which is not sensitive (not oscillating) to detectors or datasets, and it is easy to tune as demonstrated in Section V-C. Hence, NA is very useful for practical applications.

**Limitations.** One limitation of the method is the $k$-NN graph. Some neighbors can be far away, and simple averaging may not be the best solution. Possible alternatives could be weighted average and medoid. Different neighbor graphs could also be used. Some alternatives include mutual neighborhood [36], $k$-MST [37], and XNN [38]. Nevertheless, NA is already successful and we leave these ideas for future work.

The method also has the same limitation as other distance-based methods—its performance starts to degrade when the dimensions are large, as shown in the 269-dimensional Arrhythmia dataset. Neighborhood averaging still improved but the performance started to degrade if the method was iterated more than once. Such problems are common for distance-based pattern recognition methods operating in the raw attribute space. This is often referred to as the "curse of dimensionality."

## VI. Conclusions

A novel post-processing technique called neighborhood averaging (NA) technique is proposed. The technique can be used to improve any existing outlier detector. Simulations showed that it significantly improved all 12 tested outlier detectors from 0.70 to 0.79 AUC on average.

The technique does not require any complicated parameter tuning and $k$ is the only parameter. When used with a $k$-NN based baseline detector, we do not need to recalculate the $k$-NN and used the existing one with the same $k$ value as the detector. With non-$k$-NN-based detectors, setting the value of $k = 100$ was shown to provide good results for almost all datasets. It is worth noting that once NA is applied, even a poorly performing outlier detector became competitive. This can help practitioners as they have one less design component to consider.

Outlier detection is an important topic in data mining. In addition to its ability to detect outliers in static data, it can also handle dynamic cases such as time series and is therefore useful for applications like audio and video content analysis. In general, whenever *similarity* between objects can be properly predefined, whether static or dynamic, the concept of *neighborhood* can be applied. Therefore, the proposed NA can be applied to enhance performance consistently and significantly. Neighborhood averaging has the potential to be widely adopted in a variety of applications in data mining and beyond.

## REFERENCES

[1] C.C. Aggarwal, *Outlier Analysis Second Edition*, Springer International Publishing, 2016.

[2] Yang, J.; Rahardja, S.; Rahardja, S. Click fraud detection: HK-index for feature extraction from variable-length time series of user behavior. *In Proceedings of the Machine Learning for Signal Processing*, Xi'an, China, 22–24 August 2022.

[3] R. Nawaratne, D. Alahakoon, D. De Silva and X. Yu, Spatiotemporal Anomaly Detection Using Deep Learning for Real-Time Video Surveillance, in *IEEE Transactions on Industrial Informatics*, 16(1), 393-402, 2020.

[4] J. Yang, G. I. Choudhary, S. Rahardja and P. Franti, Classification of Interbeat Interval Time-series Using Attention Entropy, in *IEEE Transactions on Affective Computing*, doi: 10.1109/TAFFC.2020.3031004, 2021.

[5] J.W. Yang, R. Mariescu-Istodor, P. Fränti, Three Rapid Methods for Averaging GPS Segments, *Applied Sciences*, 9(22), 4899, 2019.

[6] Yang J, Tan X, Rahardja S. MiPo: How to Detect Trajectory Outliers with Tabular Outlier Detectors. *Remote Sensing*. 14(21):5394, 2022.

[7] J.W. Yang, S. Rahardja and P. Fränti, Mean-shift outlier detection and filtering, *Pattern Recognition*, 115, 107874, 2021.

[8] P. Fränti and J.W. Yang, Medoid-shift noise removal to improve clustering, *Int. Conf. Artificial Intelligence and Soft Computing* (ICAISC), Zakopane, Poland, 604-614, June 2018.

[9] Y. Djenouri, G. Srivastava and J. C. -W. Lin, Fast and Accurate Convolution Neural Network for Detecting Manufacturing Data, in *IEEE Transactions on Industrial Informatics*, vol. 17, no. 4, pp. 2947-2955, April 2021.

[10] K. Huang, Y. Wu, C. Wang, Y. Xie, C. Yang and W. Gui, A Projective and Discriminative Dictionary Learning for High-Dimensional Process Monitoring With Industrial Applications, in *IEEE Transactions on Industrial Informatics*, 17(1), 558-568, 2021.

[11] J.W. Yang, S. Rahardja, and P. Fränti, Outlier detection: how to threshold outlier scores, *International Conference on Artificial Intelligence, Information Processing and Cloud Computing* (AIIPCC2019), 2019.

[12] S. Ramaswamy, R. Rastogi, and K. Shim, Efficient algorithms for mining outliers from large data sets, *ACM SIGMOD Record*, 29 (2), 427-438, 2000.

[13] V. Hautamäki, I. Kärkkäinen, and P. Fränti, Outlier detection using k-nearest neighbor graph, *Int. Conf. on Pattern Recognition* (ICPR), 430-433, 2004.

[14] E.M. Knorr and R.T. Ng, Algorithms for mining distance-based outliers in large datasets, *Int. Conf. Very Large Data Bases*, 392-403, 1998.

[15] S. Shekhar, C. Lu, P. Zhang. A unified approach to detecting spatial outliers. *GeoInformatica*, 7(2), 139–166, 2003.

[16] X. Li, J. Lv, and Z. Yi, An efficient representation-based method for boundary point and outlier detection, *IEEE Trans. on Neural Networks and Learning Systems*, 29 (1), 51-62, 2018.

[17] J.W. Yang, S. Rahardja, and P. Fränti, Mean-shift outlier detection, *Int. Conf. Fuzzy Systems and Data Mining* (FSDM), 208-215, 2018.

[18] P. Fränti and J.W. Yang, Medoid-shift noise removal to improve clustering, *Int. Conf. Art. Int. Soft Computing*, 604-614, 2018.

[19] M.M. Breunig, H. Kriegel, R.T. Ng, and J. Sander, LOF: Identifying density-based local outliers, *ACM SIGMOD Int. Conf. on Management of Data*, 29 (2), 93-104, 2000.

[20] W. Hu, J. Gao, B. Li, O. Wu, J. Du and S. Maybank, Anomaly Detection Using Local Kernel Density Estimation and Context-Based Regression, in *IEEE Transactions on Knowledge and Data Engineering*, 32(2), 218-233, 2020.

[21] G.O. Campos, A. Zimek, J. Sander, R.J.G.B. Campello, B. Micenkova, E. Schubert, I. Assent, and M.E. Houle, On the evaluation of unsupervised outlier detection: measures, datasets, and an empirical study, *Data Mining and Knowledge Discovery*, 30 (4), 891–927, 2016.

[22] P.J. Rousseeuw, Least median of squares regression, *J. Am Stat Ass*, 871-880, 1984.

[23] F. Liu, T. Ting, K. Ming, and ZH. Zhou, Isolation-based anomaly detection, *ACM Transactions on Knowledge Discovery from Data* (TKDD), 6 (1), 3:1-3:39, 2012.

[24] Tan, X.; Yang, J.; Rahardja, S. Sparse random projection isolation forest for outlier detection. *Pattern Recognit. Lett.*, 163, 65–73, 2022.

[25] B. Schölkopf, J. Platt, J. Shawe-Taylor, A. Smola, and R. Williamson, Estimating the support of a high-dimensional distribution, *Neural computation*, 13 (7), 1443-1471, 2001.

[26] M-L. Shyu, S-C. Chen, K. Sarinnapakorn, and LW. Chang, A Novel Anomaly Detection Scheme based on Principal Component Classifier, *ICDM Foundation and New Direction of Data Mining workshop*, 172-179, 2003.

[27] H. Kriegel, and M. Schubert, and A. Zimek, Angle-based Outlier Detection in High-dimensional Data, *the 14th ACM SIGKDD International Conference on Knowledge Discovery and Data Mining*(KDD), 444-452, 2008.

[28] Y. Liu, et al. Generative adversarial active learning for unsupervised outlier detection, *IEEE Transactions on Knowledge and Data Engineering*, 32( 8), 1517-1528, 2019.

[29] Z. Li, et al. COPOD: Copula-Based Outlier Detection. *IEEE International Conference on Data Mining* (ICDM), 1118-1123, 2020.

[30] Z. Li, Y. Zhao, X. Hu, N. Botta, C. Ionescu and G. Chen, "ECOD: Unsupervised Outlier Detection Using Empirical Cumulative Distribution Functions," in *IEEE Transactions on Knowledge and Data Engineering*, doi: 10.1109/TKDE.2022.3159580,2022.

[31] C. C. Aggarwal, *Recommender Systems*, Springer, 268-282, 2016.

[32] Comaniciu, D., Meer, P.: Mean shift: a robust approach toward feature space analysis. *IEEE Trans. Pattern Anal. Mach. Intell.* 24(5), 603–619, 2002.

[33] P. Fränti and S. Sieranoja, K-means properties on six clustering benchmark datasets, *Applied Intelligence*, 48 (12), 4743-4759, 2018.

[34] W. Ke, J. Wei, N. Xiong, and Q. Hou. GSS: A group similarity system based on unsupervised outlier detection for big data computing. *Information Sciences*, 620, 1-15, 2023.

[35] J. Yang, Y. Chen, and S. Rahardja. Neighborhood Representative for Improving Outlier Detectors. *Information Sciences*, 625, 192-205,2023.

[36]   K.C. Gowda and G. Krishna, G. The condensed nearest neighbor rule using the concept of mutual nearest neighborhood. *IEEE Transactions on Information Theory*, *25*(4), 488-490, 1979.

[37]   C. Zhong, D. Miao, D and R. Wang, A graph-theoretical clustering method based on two rounds of minimum spanning trees. *Pattern Recognition*, *43*(3), 752-766, 2010.

[38]   P. Fränti, R. Mariescu-Istodor and C. Zhong, XNN graph. In *Joint IAPR International Workshops on Statistical Techniques in Pattern Recognition (SPR) and Structural and Syntactic Pattern Recognition (SSPR)* (pp. 207-217). Springer, Cham, 2016.